# Chases and Escapes, and Optimization Problems

Toru Ohira

Nagoya University, Japan
(Tel: 81-52-789-2824, Fax: 81-952-789-2829)

ohira@math.nagoya-u.ac.jp

**Abstract:** We propose a new approach for solving combinatorial optimization problem by utilizing the mechanism of chases and escapes, which has a long history in mathematics. In addition to the well-used steepest descent and neighboring search, we perform a chase and escape game on the "landscape" of the cost function. We have created a concrete algorithm for the Traveling Salesman Problem. Our preliminary test indicates a possibility that this new fusion of chases and escapes problem into combinatorial optimization search is fruitful.

## 1 INTRODUCTION

Problems of chasing and evading have attracted many mathematical minds in history [1]. The standard problem is to find a path of a single chaser who is trying to catch up to a single evader. One of the earliest problems is to find the path of the chaser who is chasing an evader moving in a circle with a constant speed. The condition is that the chaser also moves with a constant speed with its velocity vector pointing to the position of the evader. Unfortunately, one cannot solve this problem analytically. However, asymptotic path of the chaser is proved to be also circular (Fig.1).

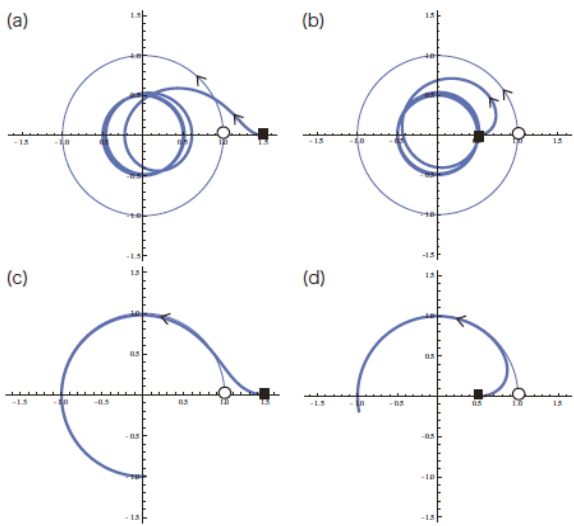

**Fig. 1.** Chase and Escapes in Circle: The white circle is the target moving in a circular path at constant speed, while black square is the target. The ratio of the speed between the two are (a,b) 1:2 (never catch up), (c,d) 1:0.95 (catch).

Interests in these problems have grown in various directions. One area is a combination with the game theory, resulting in a field called "differential game theory" [2]. Another area is called "discrete search games" [3], where a pursuer searches for an evader who is hiding behind one of the discrete number of sites. Recently, it is also extended to chases and escapes in groups, called "Group Chase and Escape" [4].

The main theme of this paper is to propose an idea of applying this "chase and escape" problem to optimization problems, such as combinatorial optimizations. There are various ways by which we can seek such applications. One way is that we consider the optimized state as an evader, and a chaser will seek that state. A slightly different approach, which we propose here, is that we perform a chase and escape game on the "landscape" of the cost function trying to find the optimized state with the lowest cost.

## 2 DESIGN OF ALGORITHM

Against the overall design concepts described in the previous section, we would like to describe one of our algorithms here. We will prepare initial two states, one is the chaser the other is the evader. Then, we change these two states according to the following rules. The evader updates its state so that the cost function becomes lower. Typically, this is done by randomly choosing a local portion of the state and change it if the new state has the lower cost, but do not change it otherwise. Thus, the evader's update rule is aimed at minimizing the cost function. The chaser, on the other hand, does not take into account of the cost function, and updates its state so that it becomes "closer" to the state of the evader in some distance measure in the state space. So, the chaser has a possibility to update to a state that has higher cost.

We will add in two more rules. The first one is to the change of roles. If during the above updating sequence, the chaser reaches a state with lower cost. Then, the roles of chaser and evader are exchanged. In other words, the chaser is always designated to the state that has a lower cost function between the two. They will keep the basic updating rule above even after the switch.

The second rule is the repulsive action when it catches up to the evader. At this stage, the states of the two become identical. We will add in this rule to separate their states. One of them will change its state and update it to search for the lower costs around its neighboring states. Then, we will designate the chaser and the evader between the two states by their associated costs.

Thus, our algorithm can be viewed as an extension of the steepest descent algorithm with neighboring search around local minima. The extension is the added mechanism of chases and escapes. It is our speculation that having a chaser whose movement is different from that of steepest descent may help "breaking out" of local minima (Fig. 2).

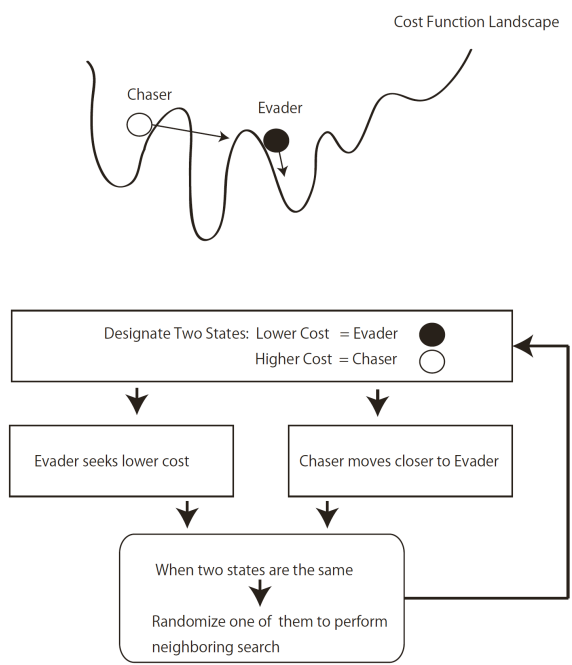

**Fig. 2.** Overall Design of the Chase and Escape mechanism for optimization problems.

## 3 APPLICATION TO TRAVELING SALESMAN PROBLEM

We here give a concrete example of our algorithm applied to a "Traveling Salesman Problem", which is a representative combinatorial optimization problem [5]. There are N cities on a two dimensional map. The problem requests to find the shortest path to visit all cities, each with only once. Though seemingly simple, this problem is one of the most difficult combinatorial optimization problems.

If we name the cities by integers (1, 2, 3, … N), then its permutation represents a path with associated distance which we identify as the "cost". The task is to find the optimal state out of N! permutations.

Let us describe our algorithm. We generate two initial states and designate the one with the shorter path as the evader and the other as the chaser. Then, for the evader, we randomly pick two cities and exchange them in its permutation. If this exchange produces the state with shorter path, we update the evader's state with the new one. Otherwise, we keep the original. Now, for the chaser, we randomly pick a city and update the state by exchanging two cities so that the city picked is now placed at the same position in the permutation as the evader.

Then, we compare the paths of the evader and chaser. If the chaser now has a shorter path, then the roles are switched. By this iteration, these two are expected to move toward a state with shorter path and approaching to each other.

When two states become the same, we take it as an indication of reaching to a local minimum and add in the next step. We randomly pick R cities and permute them. This is likely to produce a new state neighboring to the local minimum by the difference of R cities. We update this state toward the one with lower path by the two-cities exchanges. If that produces the better state, we designate it as the evader, and the original state as the chaser. Otherwise, the roles are switched, and the whole process is repeated.

## 4 SIMULATION RESULTS

We have tested our algorithm to the 52 cities Traveling Salesman Problem, which is available on the Web[6] with known shortest path of length approximately 7,544 (Figure 3).

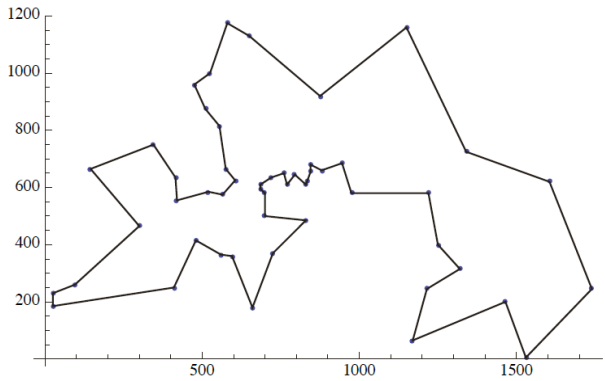

**Fig. 3.** Traveling Salesman Problem with 52 cities. The line is the shortest path. This best path length is approximately 7,544.

Out of 52! ~ $8 \times 10^{67}$ permutations, we set our parameters so that it samples at most (A) $8 \times 10^7$ and (B) $6 \times 10^8$ permutations. The parameter for the neighboring search is set as R = 3. We have shown a representative example as a result of our algorithm (Fig. 4).

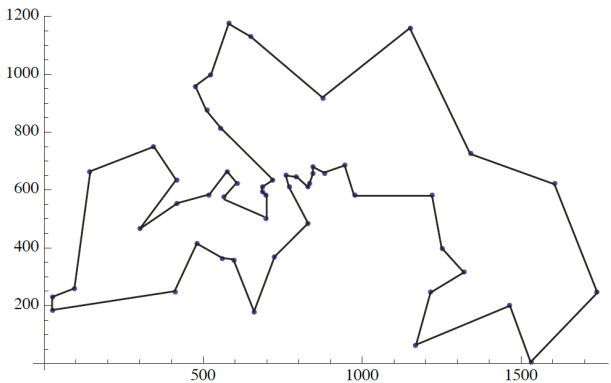

**Fig. 4.** An example of path for the Traveling Salesman Problem with 52 cities with our algorithm using chase and escape. This path length is approximately 7,940.

We repeated our trials 100 times to obtain the average path length (Ave. Path.) and the standard deviation (SDV). We compare the result (C and E) against the steepest descent with neighboring search algorithm (Simple), which takes a comparable time (sec) on the same computer. The result is summarized in Tables 1 and 2. Though marginal, the results indicates a possibility of improvements with a incorporation of the chase and escape mechanism.

**Table 1.** Comparison (A)

|  | Time(sec) | Ave. Path | SDV |
|---|---|---|---|
| Simple | 17,930 | 8,912 | 411 |
| C and E | 16,230 | 8,777 | 401 |

**Table 2.** Comparison (B)

|  | Time(sec) | Ave. Path | SDV |
|---|---|---|---|
| Simple | 81,635 | 8,401 | 310 |
| C and E | 81,871 | 8,223 | 257 |

## 5 CONCLUSION

We would like to address a couple of points based on this preliminary work of applying chases and escapes to optimization problems.

First, even though the improvement on the Traveling Salesman Problem was small, there is an indication that chases and escapes mechanism may help. Searching for optimization problems to have more notable effect is yet to be done.

Second, one can easily make this algorithm in parallel. For example, we can change the algorithm so that there is a single evader and multiple chasers. It is of interest to see how such extension can show improvements over simple parallelization without chases and escapes.

The author would like to thank Profs. Tomoaki Nogawa and Tadaaki Hosaka for fruitful discussions. This research has been supported by The Kayamori Foundation of Informational Science Advancement #K25-XVII 422.

## REFERENCES

[1] Nahin PJ (2007) Chases and Escapes: The mathematics of pursuit and evasion. Princeton University Press, Princeton
[2] Isaacs R (1965) Differential Games. Wiley, New York
[3] Rucle WH (1991) A Discrete Search Game. In: Raghavan TES et al., (eds), Stochastic Games and Related Topics, Kulwer Academic pp. 29-43
[4] Kamimura A, Ohira T (2010) Group Chase and Escape, New Journal of Physics, 12, 053013
[5] Cook JW (2012), In pursuit of the traveling salesman: mathematics at the limits of computation. Princeton University Press, Princeton
[6] The following web site has a collection of Travelling Salesman Problem. We have used one with 52cities, named in the list as *berlin52.tsp* and *berlin52.opt.tour*. http://elib.zib.de/pub/Packages/mp-testdata/tsp/tsplib/tsp/index.html